\documentclass[11pt,a4paper]{article}
\usepackage[hyperref]{acl2018}
\usepackage{times}
\usepackage{latexsym}
\usepackage{amsmath}
\usepackage{multirow}
\usepackage{url}
\usepackage{CJK,algorithm,algorithmic,amssymb,amsmath,array,epsfig,graphics,multirow,array,float,subfigure,verbatim,epstopdf}
\usepackage{enumitem}
\usepackage{color}
\usepackage{hhline}
\usepackage{booktabs}
\usepackage{flushend}

\aclfinalcopy % Uncomment this line for the final submission
\def\aclpaperid{452} %  Enter the acl Paper ID here

\title{Multi-lingual Common Semantic Space Construction via Cluster-consistent Word Embedding}
\author{Lifu Huang$^{1}$, \ \ Kyunghyun Cho$^{2}$, \ \ Boliang Zhang$^{1}$, \ \ Heng Ji$^{1}$, \ \ Kevin Knight$^{3}$ \\
  $^{1}$ Rensselaer Polytechnic Institute, $^2$ New York University \\
  $^3$ Information Sciences Institute, University of Southern California \\
  {\tt $^1$\{huangl7, zhangb8, jih\}@rpi.edu} \\
  {\tt $^2$kyunghyun.cho@nyu.edu}, \ \
  {\tt $^3$knight@isi.edu} \\}
  
\date{}

\begin{document}
\maketitle

\begin{abstract}
%that can then be processed by deeper layers of our neural CLIR networks. a
% bold, approach that will leverage a diverse set of resources to construct a single
% common semantic representation that spans thousands of languages.
%The goal of our work is to 

%\lifu{Should we say, we also regard each single token as a cluster. Thus we have four types of clusters and we construct the common space based on four types of clusters} \heng{no}

%We hypothesize that each word belongs to some certain implicit clusters in a continuous common space based on its neighbors or linguistic properties; and these word clusters should be consistently distributed across multiple languages.  

We construct a multilingual common semantic space based on distributional semantics, where words from multiple languages are projected into a shared space to enable knowledge and resource transfer across languages. 
Beyond word alignment, we introduce multiple cluster-level alignments and enforce the word clusters to be consistently distributed across multiple languages. We exploit three signals for clustering: (1) neighbor words in the monolingual word embedding space; (2) character-level information; and (3) linguistic properties (e.g., apposition, locative suffix) derived from linguistic structure knowledge bases available for thousands of languages. We introduce a new cluster-consistent correlational neural network to construct the common semantic space by aligning words as well as clusters. Intrinsic evaluation on monolingual and multilingual QVEC tasks shows our approach achieves significantly higher correlation with linguistic features than state-of-the-art multi-lingual embedding learning methods do. Using low-resource language name tagging as a case study for extrinsic evaluation, our approach achieves up to 24.5\% absolute F-score gain over the state of the art. 

% such as the Unicode Common Locale Data Repository. 

%significantly outperform state-of-the-art approaches. 
%by obtaining higher correlation and coverage scores. 
%Without using any training data for the target low-resource language, our framework achieves up to 34.1\% name tagging F-score by transferring knowledge and resources from other languages. 

%obtaining \heng{XXX}\% and \heng{XXX}\% absolute gains on Machine Translation and Name Tagging respectively. \heng{highlight the great tigrinya results}

%Experiments on multiple evaluation tasks, such as cross-lingual word similarity, word translation demonstrate the effectiveness of the two additional alignments. We also investigate the effectiveness of such common semantic space on knowledge transfer for low resource language name tagging.

%The goal of our work is to construct a multi-lingual common semantic space. The basic idea is to project multiple monolingual embeddings to a virtual common semantic space through a multi-lingual auto-encoder framework, and then reconstruct monolingual embeddings for each language by borrowing knowledge from all other languages that share semantic space. We have investigated four types of loss functions to minimize mono-lingual reconstruction errors, cross-lingual reconstruction errors, cross-lingual alignment errors and mono-lingual distribution errors respectively. Preliminary results on semantic similarity computation for 59 low-resource languages advance state-of-the-art.

\end{abstract}
\section{Introduction}
\label{sec:introduction}

%\lifu{Advances of cross-lingual embeddings}

%There are about 7,099 known living languages in the world, and 
More than 3,000 languages have electronic record, e.g., at least a portion of the Christian Bible had been translated into 2,508 different languages. %\footnote{https://www.ethnologue.com} 
However, the training data for mainstream natural language processing (NLP) tasks such as information extraction and machine translation is only available for dozens of dominant languages. In this paper we aim to construct a multilingual common semantic space where words in multiple languages are mapped into a distributed, language-agnostic semantic continuous space, so that resources and knowledge can be shared across languages. %Multilingual word embeddings, which represent words from multiple languages within one semantic space, can meet this goal. 

%\lifu{Current Techniques of crosslingual embeddings and the remaining challenges}

% Most prior work on building multilingual word embeddings either jointly learn representations for multiple languages~\cite{luong2015bilingual,duong2016learning,ammar2016massively} or first learn monolingual word embedding separately and then project them into a shared semantic space~\cite{ap2014autoencoder,duong2017multilingual}. Most previous work use bilingual word or sentence alignment as bridge to learn multilingual word representations. 
% %The only bridge that previous work has used to learn multilingual word representations is bilingual word or sentence alignment. H
% However, for extremely low resource languages, such as Oromo and Chechen, these alignments are extremely limited or unavailable, which hinders the construction of common semantic space and knowledge transfer from high resource languages to such low resource languages \heng{the above description is very verbose, need to cut down}. 

%Fortunately words 
Words can be clustered through explicit (e.g., sharing affixes of certain linguistic functions) or implicit clues (e.g., sharing neighbors from monolingual word embedding). We hypothesize that the distribution of such clusters should be consistent across multiple languages. We achieve this cluster-level consistency by aligning word clusters across languages. 
%Besides word-level, we can also align words across languages on cluster-level. 
Based on this intuition we propose to create clusters through three kinds of signals as follows, without any extra human annotation effort. Then we aggregate the embedding vectors of words in each cluster and ensure that the clusters (or the words therein) are consistent across multiple languages.

\textbf{Neighbor based clustering and alignment}. We build our common space based on correlational neural network (CorrNet) which is %commonly used to learn word representations for multiple views or languages. CorrNet is 
an extension of autoencoder framework by enabling cross-lingual reconstruction. In contrast to previous work~\cite{chandar2016correlational,rajendran2015bridge}, we extend CorrNet to \textit{neighbor-consistent correlation network} by using each word's neighbors (the nearest words within monolingual semantic space) to 
%enrich its representation in the common semantic space. \heng{give an example here}
ensure that the cross-lingual mapping from and to the common semantic space is locally smooth. For instance, the neighboring words of \textit{China} in English (\textit{Japan}, \textit{India} and \textit{Taiwan}) should be close to the neighboring words of \textit{Cina} in Italian (\textit{Beijing}, \textit{Korea}, \textit{Japan}) in the common semantic space. In other words, we encourage the consistency of neighborhoods across multiple languages. 

\textbf{Character based clustering and alignment}. 
%We introduce two additional levels of alignment for building common semantic space without any extra human annotation effort. First, we observe that 
Many related languages share very similar character set, and many words that refer to the same concept share similar compositional characters or patterns, e.g., \textit{China} (English), \textit{Kina} (Danish), and \textit{Cina} (Italian). 

\textbf{Linguistic property based clustering and alignment}. %Second, besides word-level alignment, 
Many languages also share linguistic properties, e.g., apposition, conjunction, and plural suffix (English (\textit{-s} / \textit{-es}), Turkish (\textit{-lar} / \textit{-ler}), Somali (\textit{-o})). Linguists have created a wide variety of linguistic property knowledge bases, which are readily available for thousands of languages. For example, the CLDR (Unicode Common Locale Data Repository) includes closed word classes and affixes indicating various linguistic properties. We propose to take advantage of these language-universal resources to create clusters, where the words within one cluster share the same linguistic property, and build alignment between clusters for common semantic space construction.
%words based on their linguistic properties. %For example, a Turkish word \textit{karak\"oy\textbf{de} (in Karak\"oy)} might not exist in Turkish-English word dictionaries. But from the linguistic property knowledge bases we know that the postpositions \textit{de} indicates locative suffix, and thus we can align this word with the cluster of all location names in English. 

%We demonstrate the effectiveness of 
%We evaluate our approach on multiple multilingual intrinsic evaluation tasks~\cite{}, %such as cross-lingual word similarity, multilingual word translation. Finally, we further evaluate the common semantic representations 
We evaluate our approach on monolingual and multilingual QVEC~\cite{tsvetkov2015evaluation} tasks, as well as an extrinsic evaluation on name tagging for low-resource languages. Experiments demonstrate that our framework is effective at capturing linguistic properties and significantly outperforms state-of-the-art multi-lingual embedding learning methods. 

\section{Related Work}
\label{sec:related}

%\heng{merge this paragraph with the following ones} 
Multilingual word embeddings have advanced many multilingual natural language processing tasks, such as machine translation~\cite{zou2013bilingual,mikolov2013exploiting,madhyastha2017learning}, dependency parsing~\cite{guo2015cross,ammar2016many}, and name tagging~\cite{zhangrpi,tsai2016cross}. % \heng{re-write the following sentence, not clear what you mean}
%These embeddings can provide implicit clues for the alignment of multilingual words, enabling knowledge transfer between languages when they are used as the feature representations for downstream tasks, especially in low resource settings.
%Most bilingual and multilingual word embeddings are learned using bilingual word alignment. These methods 
%The methods of learning multilingual word embeddings 
%The multilingual embedding methods can be divided into two branches: projecting multiple monolingual word embeddings into a shared semantic space 
%Most bilingual and multilingual word embeddings are learned using bilingual word alignment. These methods 
Using bilingual aligned words, previous methods project multiple monolingual embeddings into a shared semantic space using linear mappings~\cite{mikolov2013exploiting,rothe2016ultradense,zhang2016ten,marcobaroni2015hubness,xing2015normalized} %, %Among which,~\newcite{xing2015normalized} and~\newcite{smith2017offline} showed that adding an orthogonality
%constraint to the mapping can significantly improve performance, and has a closed-form
%solution. However, such solutions can only be applied to project individual pair of languages. 
%The second is based on 
or canonical correlation analysis (CCA)~\cite{ammar2016massively,faruqui2014improving,lu2015deep}. Compared with CCA, which only optimizes the correlation for each individual pair of languages, linear mapping based methods can jointly optimize all the languages in the common semantic space. %Correlational neural networks (CorrNets)~\cite{chandar2016correlational,rajendran2015bridge} extends autoencoder framework by enabling cross-lingual reconstruction. %It has been demonstrated to be effective in learning multi-view representations. In this work, 
We focus on learning linear mappings to construct the common semantic space and adopt correlational neural networks~\cite{chandar2016correlational,rajendran2015bridge} as the basic model. %Compared with 
In contrast to previous work which only exploited monolingual word semantics, we introduce multiple cluster-level alignments.%monolingual context and character compositional semantics to further improve the semantic representation of each word. 

Beyond word alignment, another branch of approaches for multilingual word embeddings are based on parallel or comparable data, such as parallel sentences~\cite{ap2014autoencoder,gouws2015bilbowa,luong2015bilingual,hermann2014multilingual,schwenk2017learning}, phrase translations~\cite{duong2016learning} and comparable documents~\cite{vulic2015bilingual}. Moreover, to reduce the need of bilingual alignment, several approaches have been designed to learn cross-lingual embeddings based on a small seed dictionary~\cite{vulic2016role,zhang2016ten,artetxe2017learning}, 
%However, such methods cannot be extended to learn high-quality multilingual word embeddings.
%~\newcite{heyman2017bilingual} also incorporate character-level information for bilingual lexicon induction and demonstrate its effectiveness.  
%\heng{add all of these (1) distributed monolingual representations on character-level~\cite{Costa-jussa2016,Luong2016}, or subword-level  ~\cite{Salam2012,Rei2016,Sennrich2016,Yang2017}, or bi-lingual word embeddings~\cite{Madhyastha2017}.
%add all papers in EMNLP17 tutorial on multilingual embeddings} However, ..\heng{make this complete}
%Finally, there have been a few attempts to align monolingual word vector spaces 
or even with no supervision~\cite{cao2016distribution,zhang2017earth,zhang2017adversarial,conneau2017word}. However, such methods are still limited to bilingual word embedding learning and remaining to be explored for common semantic space construction.

\section{Approach}
%\label{sec:approach}

%In this section, we first briefly introduce the approach of AutoEncoders for monolingual semantic space projection, then provide more detailed information of our proposed multilingual-AuenEncoders framework for the common semantic space construction.
%\input{3.1ae}
%\input{3.2multilingualAE}
%\input{3.3linguistic}

%%%%%%%%%%%%%%%%%%%%%%%%%%%%%  Old  %%%%%%%%%%%%%%%%%%%%%%%%%%%%%
%\input{3.1overview}
%\input{3.2monolingual}
%\input{3.3alignment}

%%%%%%%%%%%%%%%%%%%%%%%%%%%%%  New  %%%%%%%%%%%%%%%%%%%%%%%%%%%%%
\subsection{Overview}

%\heng{1. the colors are confusing, explain what they mean; 2. 'monolingual crosslingual reconstruciton' is also confusing, re-phrase it; 3. you still need a more complicated picture like those in the slides to show the novel parts, like context-augmented, structure-level alignment, character-level, etc}

%\heng{to improve figure: 1. make it clear which parts are neighbor words; 2. add property alignment; 3. add labels about multiple levels of alignment; 4. maybe use a rare word as example instead of China}

\begin{figure*}[!htb]
\centering
\includegraphics[width=.98\textwidth]{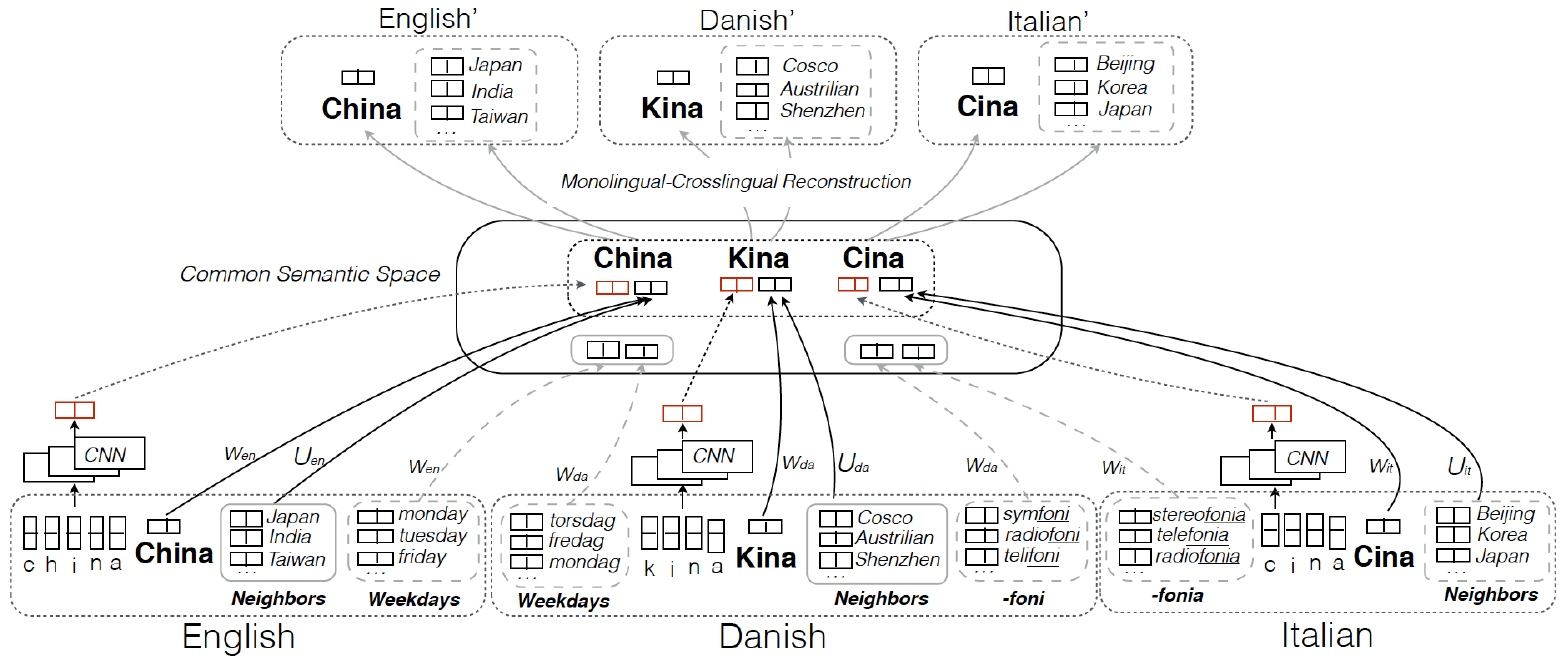}
\caption{Architecture Overview. In each monolingual semantic space, the words within solid rectangle denote a neighbor based cluster and the words within dotted rectangle denote a linguistic property based cluster.}
\label{overview}
\end{figure*}

Figure~\ref{overview} shows the overview of our neural architecture. 
%\heng{we need something like the following:  The goal of our multilingual multi-level autoencoder is, for each monolingual embedding for each language, to automatically learn a reduced dimensional vector representation, then to reconstruct the input from a new vector from the shared common space with minimal error.
% We assume a projected semantic space will better capture monolingual properties if we can reconstruct the original monolingual semantic space from it and the distributions of words in the original semantic space is consistent with their distributions in the projected
% semantic space. Therefore we plan to minimize the following three loss functions: (1) Monolingual
% reconstruction errors: project from monolingual embedding space to common semantic space, then
% reconstruct this monolingual embedding space; (2) Cross-lingual reconstruction errors: project from
% common semantic space to monolingual embedding space, using aligned words from other languages;
% and (3) Cross-lingual alignment errors: minimize the distance of the aligned word representations in
% the common semantic space.}
%\lifu{Rewrite till here}
We project all monolingual word embeddings into a common semantic space based on word-level as well as cluster-level alignments and learn the transformation functions. 
%multi-level alignments and optimize the transformation function. 
First, on word-level, 
we build a %context augmented 
neighborhood-consistent CorrNet to %exploit %incorporate %the monolingual ``context'' information into 
%the set of neighboring words 
augment word representations with neighbor based clusters and align them in the common semantic space. In addition, we apply a language-independent convolutional neural networks to compose character-level word representation and concatenate it with word representation in the common semantic space. Finally, we construct clusters based on linguistic properties, such as closed word classes and affixes, %to induce linguistic property level clusters 
and align them in the common semantic space. We jointly optimize for all the alignments in the common semantic space for each pair of languages.

%also incorporate the structure-level alignment. For example, in English, ``-s'' is a plural suffix which is shared by many word pairs, e.g., $<$book, books$>$; and in Uzbek, ``-lar'' is a plural suffix. Therefore we can induce a vector representation for each linguistic property, and align these property representations in the common semantic space. 

% adopt a character-aware neural language model to learn monolingual word embeddings: Each word is represented as a sequence of characters, while all languages share the same set of characters and the same character lookup embeddings. We apply a Convolutional Neural Network (CNN) over each sequence of characters, and a max-over-time pooling function to get the character aware word representations. All word representations will be further optimized by a multi-layer LSTM and a softmax function, and minimize the loss between the predicted distribution over next word and the actual next word.

% The second step is to project all monolingual word embeddings into a common semantic space based on multi-level alignments. For each language, we are aiming to learn an optimized transformation function. Besides lexical level alignment based on bilingual dictionaries, we also incorporate the structure-level alignment. For example, in English, ``-s'' is a plural suffix which is shared by many word pairs, e.g., $<$book, books$>$, while in Uzbek, ``-lar'' is a plural suffix. We can induce a vector representation for each linguistic feature, and align these feature representations in the common semantic space. 
\subsection{Basic Model}
\label{sec:corrnet}
We briefly describe the basic model for learning the common semantic space: correlational neural networks (CorrNets)~\cite{chandar2016correlational,rajendran2015bridge}. CorrNets have been widely adopted for learning multilingual or multi-view representations. %Figure~\ref{corrnet} shows the basic architecture of a CorrNet. 
It combines the advantages of canonical correlation analysis (CCA) and autoencoder (AE). 

% \begin{figure}[!htb]
% \centering
% \includegraphics[width=.35\textwidth]{Figures/corrnet.pdf}
% \caption{CorrNet for Learning Multilingual Word Embeddings}
% \label{corrnet}
% \end{figure}

%\heng{before you explain these formulas you should say you are using English and Danish as an example language pair}

%We take English and Danish as an example to show the construction of the common semantic space. 
Given the bilingual aligned word pairs between two languages $l_1$ and $l_2$, we first use their monolingual word embeddings to initialize each word with a vector and obtain
$M_{l_1}\in\mathbb{R}^{|V_{l_1}|\times d_{l_1}}$ and $M_{l_2}\in\mathbb{R}^{|V_{l_2}|\times d_{l_2}}$, where $V_{l_1}$ and $V_{l_2}$ are the bilingual dictionary of $l_1$ and $l_2$. %the dictionary for English and Danish extracted from their bilingual dictionary, 
$V_{l_1}^{i}$ is the translation of $V_{l_2}^{i}$, and $d_{l_1}$ and $d_{l_2}$ are the vector dimensionalities. Then for each language we learn a linear projection function to project $M_{l_1}$ and $M_{l_2}$ into the common semantic space:
\begin{displaymath}
H_{l_1} = \sigma(M_{l_1}\cdot W_{l_1} + b_{l_1}) \ ,
\end{displaymath}
\begin{displaymath}
H_{l_2} = \sigma(M_{l_2}\cdot W_{l_2} + b_{l_2}) \ ,
\end{displaymath}
where $H_{l_1}\in\mathbb{R}^{V_{l_1}\times h}$ and $H_{l_2}\in\mathbb{R}^{V_{l_2}\times h}$ are the vector representations for $V_{l_1}$ and $V_{l_2}$ in the common semantic space respectively. $h$ is the vector dimensionality in the shared semantic space. 
$W_{l_1}\in\mathbb{R}^{V_{l_1}\times h}$ and $W_{l_2}\in\mathbb{R}^{V_{l_2}\times h}$ are the transformation matrices, and $b_{l_1}$ and $b_{l_2}$ are the bias vectors. $\sigma$ denotes Sigmoid function.

After we project the monolingual embeddings into the common semantic space, we further reconstruct $M_{l_1}$ and $M_{l_2}$ from $H_{l_1}$ and $H_{l_2}$ separately:
\begin{displaymath}
M^{'}_{l_1} = \sigma(H_{l_1}\cdot W^{\top}_{l_1} + b^{'}_{l_1}) \ ,
\end{displaymath}
\begin{displaymath}
M^{*}_{l_1} = \sigma(H_{l_2}\cdot W^{\top}_{l_1} + b^{'}_{l_1}) \ ,
\end{displaymath}
\begin{displaymath}
M^{'}_{l_2} = \sigma(H_{l_2}\cdot W^{\top}_{l_2} + b^{'}_{l_2}) \ ,
\end{displaymath}
\begin{displaymath}
M^{*}_{l_2} = \sigma(H_{l_1}\cdot W^{\top}_{l_2} + b^{'}_{l_2}) \ ,
\end{displaymath}
where $b^{'}_{l_1}$, $b^{'}_{l_2}$ are the bias vectors. $M^{'}_{l_1}$ and $M^{'}_{l_2}$ are the monolingual reconstructions of $M_{l_1}$ and $M_{l_2}$ from the common space, and $M^{*}_{l_1}$ and $M^{*}_{l_2}$ are cross-lingual reconstructions. $W_{l_1}^{\top}$ and $W_{l_2}^{\top}$ are the transposes of $W_{l_1}$ and $W_{l_2}$ respectively. %\heng{maybe walk through the examples}

To learn the common semantic space, we minimize the distance between the aligned word  vectors as well as 
%in the common semantic space. In addition, we also minimize 
the loss of monolingual and cross-lingual reconstruction:
% minimize the following objective functions by finding the parameters $\theta=\{W_{l}, b_l, W^{'}_l, b^{'}_l\}$, where $l$ denotes a specific language:
\begin{multline*}
O_{W} = \sum_{\{l_i, l_j\}\in A} L(M^{'}_{l_i}, M_{l_i}) + L(M^{*}_{l_i}, M_{l_i}) \\
+ L(M^{'}_{l_j}, M_{l_j}) + 
L(M^{*}_{l_j}, M_{l_j}) + L(H_{l_i}, H_{l_j}) \ ,
\end{multline*}
where $l$ denotes any specific language that we want to project into the common semantic space, $A$ denotes all bilingual dictionaries, and $L$ denotes a similarity metric. In our work, we use cosine similarity as the similarity metric.
% \subsection{\textit{Context} Augmented CorrNets}
\subsection{Neighborhood-Consistent CorrNet}
\label{subsec:context}
%\heng{sometimes CorrNets sometimes CorrNet, make these consistent throughout the paper}
CorrNet can project multiple monolingual word embeddings into a common semantic space using bilingual word alignment. However, the same concepts may have different semantic bias in various languages. 
%even when it refer to the same meaning. 
For example, the top five nearest words of the concept ``\textit{China}'' are:  (\textit{Japan, India, Taiwan, Chinese, Asia}) in English, (\textit{Cosco, Shenzhen, Australian, Shanghai,  manufacturing}) in Danish, and (\textit{Beijing, Korea, Japan, aluminum, copper}) in Italian respectively. %\heng{you had a point about showing the bis in neighbor distribution; if you still want to talk about it (only if it's useful to make your point) then you should write it clearly here; not sure if the following example helps on what you want to say here: For example, ``fire'' in English has several distinct senses, and each sense corresponds to a distinct Chinese concept.} 
%\heng{I'm very confused by this example; maybe put in real context sentences to explain what cased this difference. } 
The neighboring words can reflect the semantic meanings of each concept within each semantic space. %We see that, for the same concept, they have distinct semantic bias in different languages. 
In order to ensure the consistency of the neighborhoods within the common semantic space and make the cross-lingual mapping locally smooth, we propose to augment monolingual word representation with its top-$N$ nearest neighboring words from the original monolingual semantic space.\footnote{We set $N$ as 5 in our experiments.}
%Thus, we propose to augment monolingual word representation with contextual words when projecting it into the common semantic space. 
%Here the \textit{context} means the top-$N$\footnote{We set $N$ as 5 in our experiments.} nearest words within the original monolingual semantic space. %Inspired by previous work~\cite{}, we incorporate contextual information into the representation of each word when projecting it into the common semantic space.

%introduce a contextual gate to control how much context information should be incorporated into each word when project it into the common semantic space.

%\heng{before you start to use these abbreviations like en and da, explain what they represent}

Given the monolingual embeddings of the bilingual aligned words for two languages $l_1$ and $l_2$, $M_{l_1}$ and $M_{l_2}$, for each word, we extract the top-$N$ nearest neighbors and construct the neighborhood clusters. Each cluster $t_{l} = \{w_1, w_2, ..., w_{|t_{l}|}\}$ in language $l$ is represented by
\begin{displaymath}
c_{t_l} = \frac{1}{|t_l|}\sum_{w\in t_{l}}E_w \ ,
\end{displaymath}
where $E_w$ denotes the monolingual word embedding for $w$.

We obtain all the neighborhood cluster vector representations $C_{l_1}$, $C_{l_2}$ for $l_1$ and $l_2$. We incorporate these neighborhood cluster information into the common semantic space when projecting monolingual embeddings:
\begin{align}
\label{eq:context_proj}
H_{l_1} = \sigma(M_{l_1}\cdot W_{l_1} + C_{l_1}\cdot U_{l_1} + b_{l_1}), 
\nonumber
\\
H_{l_2} = \sigma(M_{l_2}\cdot W_{l_2} + C_{l_2}\cdot U_{l_2} + b_{l_2}),
\end{align}
%\heng{maybe just use e instead of en, and d instead of da?}
%\heng{and we need an architecture figure to show this context gate}

Besides the monolingual and cross-lingual reconstructions for $M_{l_1}$ and $M_{l_2}$ in CorrNets, we also add monolingual and cross-lingual reconstructions for the neighborhood clusters:
\begin{displaymath}
C^{'}_{l_1} = \sigma(H_{l_1}\cdot U^{\top}_{l_1} + b^{*}_{l_1}) \ ,
\end{displaymath}
\begin{displaymath}
C^{*}_{l_1} = \sigma(H_{l_2}\cdot U^{\top}_{l_1} + b^{*}_{l_1}) \ ,
\end{displaymath}
\begin{displaymath}
C^{'}_{l_2} = \sigma(H_{l_2}\cdot U^{\top}_{l_2} + b^{*}_{l_2}) \ ,
\end{displaymath}
\begin{displaymath}
C^{*}_{l_2} = \sigma(H_{l_1}\cdot U^{\top}_{l_2} + b^{*}_{l_2}) \ ,
\end{displaymath}

In addition to optimizing the loss functions described in the Section~\ref{sec:corrnet}, we further optimize the monolingual and cross-lingual reconstruction for neighborhood clusters:
\begin{multline*}
O_{N} = \sum_{\{l_i, l_j\}\in A} L(C^{'}_{l_i}, C_{l_i}) + L(C^{*}_{l_i}, C_{l_i}) \\
+ L(C^{'}_{l_j}, C_{l_j}) + 
L(C^{*}_{l_j}, C_{l_j}) \ ,
\end{multline*}

% \begin{displaymath}
% O = O_{M} + O_{C} 
% \end{displaymath}

\subsection{Character-Level Word Alignment}

Bilingual word alignment is not always enough to induce a common semantic space, especially for low-resource languages. %From our observations, 
Although the words that refer to the same concept are not exactly the same in multiple languages, they usually share a set of similar characters, especially in related languages written in the same script, such as Amharic and Tigrinya. For example, the same entity is spelled slightly differently in three languages:  \textit{Semsettin Gunaltay} in English, \textit{Şemsettin Günaltay} in Turkish, and \textit{Semsetin Ganoltey} in Somali. Beyond word-level alignment, we introduce character-level alignment by composing word representations from its compositional characters using convolutional neural networks (CNN). For each language, we adopt a language-independent CNN to generate character-level word representation. 

\paragraph{Character Lookup Embeddings} Let $S_{l}$ be the character set for language $l$ and $E_{S_l}\in\mathbb{R}^{|S_l|\times d}$ be the character lookup embeddings, where $d$ is the dimensionality of each character embedding. Here, we use a simple yet effective method to induce character embeddings from word embeddings. For each character $c$, we average the embeddings of all words which contain the character. The character embeddings will be further tuned by the model. 

\paragraph{Character-Level CNN~\cite{kim2016character}} %For each language $l$, for each word $w_{i}^{l}$, which consists of a sequence of $k$ characters $[c_1, c_2, ..., c_k]$, we train a CNN to generate the representation for $w_{i}^{l}$.
The input layer is a sequence of characters of length $k$ for each word. Each character is represented by a d-dimensional lookup embedding. Thus each input sequence is represented as a feature map of dimensionality $d\times k$.

The convolution layer is used to learn the representation for each sliding $n$-gram characters. We make $p_i$ as the concatenated embeddings of $n$ continuous columns from the input matrix, where $n$ is the filter width. We then apply the convolution weights $W\in\mathbb{R}^{d\times nd}$ to $p_i$ with a biased vector $b\in\mathbb{R}^{d}$ as follows:
\begin{displaymath}
\footnotesize
p^{'}_{i} = \tanh(W\cdot p_i + b)
\end{displaymath}

All $n$-gram representations $p_{i}^{'}$ are used to generate the word representation $y$ by max-pooling.
%taking the max value:
% \begin{displaymath}
% \footnotesize
% y_{j} = \max_{i}\{p_{i,j}^{'}\},  \quad j=1,2,...,d^{'}
% \end{displaymath}

In our experiments, we apply multiple filters with various widths to obtain the representation for word $w_{i}^{l}$. The final character-level word representation $\hat{w}_{i}^{l}$ is the concatenation of all word representations with varying filter widths.

\paragraph{Cross-lingual Mapping} Given the bilingual aligned word pairs, we directly minimize the distance of the character-level word representations in the common semantic space by:
\begin{displaymath}
O_{char} = \sum_{\{l_i, l_j\}\in A} L(\hat{W}_{l_i}^{char}, \hat{W}_{l_j}^{char})
\end{displaymath}

The final word representation of $w_{i}^{l}$ in the common semantic space is the concatenation of character-level word presentation $\hat{w}_{i}^{l}$ and projected word representation $h_{i}^{l}$.
\begin{table}[t]
\footnotesize
\centering
\begin{tabular}{l|l}
% \hline 
Class Name & Words / Word Pairs \\
\toprule
Colors & white, yellow, red, blue, green ...\\
Weekdays & monday, tuesday, friday, sunday ...\\
Months & january, february, march, april ...\\
%cardinal 
numbers & one, two, three, four, five ...\\
%ordinal numbers & first, second, third, fourth, fifth ...\\
pronouns & i, me, you, he, she, her, they ...\\
prepositions & of, in, on, for, from, about ...\\
conjunctions & but, and, so, or, when, while ...\\
clothes & hat, shirt, pants, skirt, socks ...\\
\midrule
-like & (god, godlike), (bird, birdlike) ... \\
%-worthy & \\
-able &  (accept, acceptable), (adopt, adoptable) ...\\
micro-  &  (gram, microgram), (chip, microchip) ...\\
auto- & (maker, automaker), (gas, autogas) ...\\
% pseudo- & (random, pseudorandom), (code, pseudocode) ...\\
% \hline
\end{tabular}
\caption{Examples of closed word classes and linguistic properties based clusters}
\label{clusters}
\end{table}

\subsection{Linguistic Property Alignment}
%\lifu{The main contribution of this alignment is to enrich bilingual alignment, instead of replacing bilingual word alignment}
%Bilingual word clusters based on functions: Months, Weekdays, Personal Pronoun, Colors, 
%Suffix/Prefix based clusters of word-pairs, including plural suffix ...

Linguists have made great efforts at building linguistic property knowledge bases for thousands of languages in the world. These knowledge bases include a large number of topological properties (phonological, lexical and grammatical) which we will use to build a high-level alignment between words across languages. We exploit the following resources:
\begin{itemize}
\item \textbf{CLDR} (Unicode Common Locale Data Repository)\footnote{http://cldr.unicode.org/index/charts} which includes multilingual gazetteers for months, weekdays, cardinal and ordinal numbers; 
\item \textbf{Wiktionary}\footnote{https://en.wiktionary.org} which is a multilingual, web-based collaborative project to create an English content dictionary, includes word and prefix/suffix dictionaries for 1,247 languages; 
\item \textbf{Panlex}\footnote{http://panlex.org/} database which contains 1.1 billion pairwise translations among 21 million expressions in about 10,000 language varieties. 
%Wikitionary is a multilingual, web-based collaborative project to create an English content dictionary of all words in 1,247 languages; and 
\end{itemize}

\begin{table}[t]
\small
\centering
\begin{tabular}{l|c}
% \hline 
Parameter Name & Value \\
% \hline
\toprule
Monolingual Word Embedding Size & 512 \\
Multilingual Word Embedding Size & 512 \\
\# of Filters in Convolution Layer & 20 \\
Filter Widths & 1, 2, 3 \\
Batch Size & 500 \\
Initial Learning Rate & 0.5 \\
Optimizer & Adadelta \\
% \hline
\end{tabular}
\vspace{-3mm}
\caption{Hyper-parameters.}
\vspace{-3mm}
\label{parameters}
\end{table}

\begin{table*}[t]
\footnotesize
\begin{center}
\begin{tabular}{ll|cc|cc|cc|cc}
% \hline
 & & \multicolumn{4}{c|}{3 Languages} & \multicolumn{4}{c}{12 Languages} \\ 
 \cmidrule{3-10}
 
 & & \multicolumn{2}{c|}{Monolingual}  & \multicolumn{2}{c|}{Multilingual}  & \multicolumn{2}{c|}{Monolingual}  & \multicolumn{2}{c}{Multilingual} \\ 
  \cmidrule{3-10}
  
 & & QVEC & QVEC-CCA & QVEC & QVEC-CCA & QVEC & QVEC-CCA & QVEC & QVEC-CCA\\
 \toprule
\multicolumn{2}{l|}{MultiCluster}  & 10.8 & 9.1 & 63.6 & 45.8 & 10.4 & 9.3 & 62.7 & 44.5 \\
% \hline
% multiCCA & \textbf{47.3} [50.6] & \textbf{58.2} [95.1] & 40.9 [66.5] & 10.8 [97.8] & 8.5 [75.6] & 63.8 [97.8] & 43.9 [75.6] \\
\multicolumn{2}{l|}{MultiCCA}  & 10.8  & 8.5  & 63.8  & 43.9 & 10.8 & 8.5 & 63.9 & 43.7 \\
% \hline
\multicolumn{2}{l|}{MultiSkip}  & 7.8 & 7.3 & 57.3 & 36.2 & 8.4 & 7.2 & 59.1 & 36.5 \\
%MultiTrans & 8.0 & 5.4 & 65.7 & \textbf{46.1} &  8.1  & 5.3 & 65.8 & 46.2 \\ 
\multicolumn{2}{l|}{MultiCross} & - & - & - & - &  11.9 & 8.6 & 46.4 & 31.0 \\
\midrule
%\hline
%\hline
% CorrNet$^{w}$ & 37.9 & 48.9 & 37.9 & 14.9 & - & 63.2 & 41.1  \\
\multirow{5}{*}{\rotatebox{90}{CorrNet}} & W  & 14.8 & 11.3 & 63.6 & 43.4 & 14.7 & 13.2 & 63.8 & 43.9 \\
%\hline
 & W+N  & \textbf{15.9} & 12.7 & 64.5 & 45.3 & 15.5 & 13.6 & 65.0 & 46.4 \\
%\hline
%CorrNets$^{w+char}$  & 13.7 & 10.5 & 65.1 & 42.4 & 13.9 & 11.6 & 65.7 & 43.1 \\
%\hline
 & W+N+Ch  & 15.2 & 12.1 & 66.3 & 44.5 & 14.8 & 12.9 & 67.2 & \textbf{47.3} \\
 & W+N+L  & 15.8 & \textbf{12.8} & 64.3 & 45.3 & \textbf{16.3} &  \textbf{14.5} & 65.0 & 45.9 \\
%\hline
%CorrNets$^{w+N+L}$ &  &  &  &  &  &  &  &   \\
%\hline
 & W+N+Ch+L  & 15.5 & 12.7 & \textbf{66.5} & \textbf{46.3} & 14.9 & 13.1 & \textbf{67.3} & 47.2 \\
%  \hline
\end{tabular}
\end{center}

\vspace{-3mm}
\caption{QVEC and QVEC-CCA scores. W: word alignment. N: neighbor based clustering and alignment. Ch: character based clustering and alignment. L: linguistic property based clustering and alignment.}
%Experiments on 3 Languages and 12 Languages}
\label{3lang}
\vspace{-3mm}
\end{table*}

% talk about linguistic properties such as 

%In order to enrich the bilingual alignments for multilingual embedding learning, we further introduce a new linguistic property alignment based on the following resources:.

%Two such examples are CLDR (Unicode Common Locale Data Repository)\footnote{http://cldr.unicode.org/index/charts} which includes multilingual gazetteers for months, weekdays, cardinal and ordinal numbers, and Wiktionary, which includes word and prefix/suffix dictionaries for 1,247 languages. 

%Wikitionary is a multilingual, web-based collaborative project to create an English content dictionary of all words in 1,247 languages; and Panlex database contains 1.1 billion pairwise translations among 21 million expressions in about 10,000 language varieties. 

We mainly exploit two types of linguistic properties to extract word clusters. The first type is  closed word classes, such as colors, weekdays, and months. Table~\ref{clusters} shows some examples of the word clusters we automatically extracted from CLDR and Wiktionary for English. The second type of word clusters are generated based on morphological information, including affixes that indicate various linguistic functions. These properties tend to be consistent across many languages. 
%Most languages share very similar sets of morphologies with distinct strings. 
For example, ``-like'' is a suffix denoting ``similar to'' in English, while in Danish ``-agtig'' performs the same function. For each affix, we extract a set of word pairs (\textit{basic word}, \textit{extended word with affix}) to denote its semantics from each language.

%Thus, we 
We extract a set of word clusters from each language, and align the clusters based on their functions defined in CLDR, Wiktionary and Panlex. For each language $l$, each cluster $r^{l}_i\in R^{l}$ contains a set of words or word-pairs sharing the same function. We use the average operation to obtain an overall vector representation for each cluster $M_{l}^{R}$.\footnote{For each word pair, we use the vector of the extend word minus the vector of the basic word as the vector representation of the word pair.} Then, we project the cluster-level vectors into the shared semantic space and minimize the distance between them:
\begin{displaymath}
H_{l_i}^{R} = \sigma(M_{l_i}^{R}\cdot W_{l_i} + b^{R}_{l_i}) \ ,
\end{displaymath}
\begin{displaymath}
H_{l_j}^{R} = \sigma(M_{l_j}^{R}\cdot W_{l_j} + b^{R}_{l_j}) \ ,
\end{displaymath}
\begin{displaymath}
O_{R} = \sum_{\{l_i, l_j\}\in A} L(H^{R}_{l_i}, H^{R}_{l_j}) \ ,
\end{displaymath}
where $W$ is the same as the $W$ used in Section~\ref{subsec:context} for each language. We finally optimize the sum of the losses by finding the parameters $\theta=\{W_{l}, b_l, b^{'}_l, U_l, b^{*}_l, $ CNN$_{l}, b^{R}_{l}\}$, where $l$ denotes a specific language:
\begin{displaymath}
O_{\theta} = O_{W} + O_{N} + O_{char} + O_{R}
\end{displaymath}

\section{Experiments}
\label{sec:evaluation}

\subsection{Experiment Setup}

%Following previous work~\cite{ammar2016massively,duong2017multilingual}, we first evaluate our common semantic space on several intrinsic evaluation tasks, which aim to evaluate the quality of multilingual embeddings from two aspects: (1) whether semantically similar words from the same language are close in the common semantic space (2) whether cross-lingual aligned words are semantically close in the common semantic space. 
Previous work~\cite{ammar2016massively,duong2017multilingual} evaluated multilingual word embeddings on a series of intrinsic (e.g., monolingual and cross-lingual word similarity, word translation) and extrinsic (e.g., multilingual document classification, multilingual dependency parsing) evaluation tasks. Compared with previous work, we aim at incorporating more linguistic features into the multilingual embeddings, which can be helpful for downstream NLP tasks. In order to evaluate the quality of the multilingual embeddings, we use QVEC~\cite{tsvetkov2015evaluation} tasks (details will be described in Section~\ref{section_intrinsic}) as the intrinsic evaluation platform. In addition, to demonstrate the effectiveness of our common semantic space for knowledge transfer, especially for low-resource scenarios, we adopt the low-resource language name tagging task %as a case study 
for extrinsic evaluation.% Compared with some traditional multilingual extrinsic evaluation tasks, the name tagging task requires more linguistic features, such as morphological information.
%and adopt name tagging ans the extrinsic evaluation task.

For fair comparison with state-of-the-art methods on building multi-lingual embeddings~\cite{ammar2016massively,duong2017multilingual}, we use the same monolingual data and bilingual dictionaries as in their work. 
%are the same as previous work for fair comparison.
We build multilingual word embeddings for 3 languages (\textit{English, Italian, Danish}) and 12 languages (\textit{Bulgarian, Czech, Danish, German, Greek, English, Spanish, Finnish, French, Hungarian, Italian, Swedish}) respectively. The monolingual data for each language is the combination of the Leipzig Corpora Collection\footnote{http://wortschatz.uni-leipzig.de/en/download/} and Europarl.\footnote{http://www.statmt.org/europarl/index.html} The bilingual dictionaries are the same as those used in~\newcite{ammar2016massively}.\footnote{http://128.2.220.95/multilingual/data/} 

%the testbed.

%\footnote{According to~\newcite{duong2017multilingual}, for fair comparison and with respect to the word coverage, we don't use MultiTrans as a baseline since its coverage is much lower.}

For each task, we evaluate the performance of our common semantic space in comparison with previously  published multilingual word embeddings (MultiCluster, MultiCCA, MultiSkip, and MultiCross). MultiCluster~\cite{ammar2016massively} groups multilingual words into clusters based on bilingual dictionaries and forces all the words from various languages within one cluster share the same embedding. MultiCCA~\cite{ammar2016massively,faruqui2014improving} uses CCA to estimate linear projections for each pair of languages. MultiSkip is an extension of the multilingual skip-gram model~\cite{luong2015bilingual}, which requires parallel data. MultiCross is an approach to unify bilingual word embeddings into a shared semantic space using post hoc linear transformations~\cite{duong2017multilingual}.

Table~\ref{parameters} lists the hyper-parameters used in the experiments.

\begin{table*}[t]
\small
%\scriptsize
\footnotesize
\begin{center}
\begin{tabular}{l|l|cc|cc}
% \hline
\multicolumn{2}{c|}{} &  \multicolumn{2}{c|}{QVEC} & \multicolumn{2}{c}{QVEC-CCA} \\
\multicolumn{2}{c|}{} &  Monolingual  & Multilingual  & Monolingual  & Multilingual \\
 \toprule
 \multirow{2}{1.0cm}{40,000} & multiCCA  & 10.8 & 8.5 & 63.8 & 43.9 \\
 %\cline{2-9}
& CorrNet W & 14.8 & 11.3 &  63.6 & 43.4 \\
& CorrNet W+N+Ch+L  & \textbf{15.5} & \textbf{12.7} & \textbf{66.5} & \textbf{46.3} \\
\midrule
%\hline
\multirow{2}{1.0cm}{10,000} & multiCCA  & 9.8 & 6.5 & 63.6 & 42.3 \\
 %\cline{2-9}
& CorrNet W  & 14.8 & 11.3 & 63.4 & 43.0 \\
& CorrNet W+N+Ch+L  & \textbf{15.4}  & \textbf{12.1}  & \textbf{66.4}  & \textbf{46.2} \\
%\hline
\midrule
\multirow{2}{1.0cm}{2,000} & multiCCA  & 9.9 & 6.2 & 63.6 & 40.9 \\
  %\cline{2-9}
& CorrNet W  & 14.5 & 7.1 & 62.0 & 39.2 \\
& CorrNet W+N+Ch+L & \textbf{14.7}  & \textbf{11.7}  & \textbf{66.6} & \textbf{45.5} \\
 \midrule
\end{tabular}
\vspace{-3mm}
\end{center}
\caption{Results using bilingual lexicons with varying sizes (40,000, 10,000, 2,000) and three languages. CorrNet W+N+Ch+L is the proposed approach with all the cluster types.}
\label{3lang2000}
\end{table*}

\subsection{Intrinsic Evaluation: QVEC}
\label{section_intrinsic}
%\heng{justify why you use this data set: in order to compare with previous methods}

%\heng{add justification about why we only use QVEC measures: because they focus on linguistic properties, check correlation scores}

%\heng{in each table, highlight which rows are 'our approach'}
%\heng{maybe only keep the results with the same vocabulary size}

In order to evaluate the quality of multilingual embeddings, especially on linguistic aspect, we adopt QVEC~\cite{tsvetkov2015evaluation} as the intrinsic evaluation measure. It evaluates the quality of word embeddings based on the alignment of %dimensions of 
distributional word vectors to %dimensions in the 
linguistic feature vectors extracted from manually crafted lexical resources, e.g., SemCor~\cite{miller1993semantic}. 
%Given the word vocabulary $V$, where each word is associated with a distributional word vector $x\in \mathbb{R}^{D\times 1}$ and a linguistic word vector $s\in  \mathbb{R}^{P\times 1}$, we can get the distributional word vector matrix $X\in \mathbb{R}^{D\times |V|}$ and linguistic property matrix $S\in \mathbb{R}^{P\times |V|}$. The alignment matrix $A\in \{0, 1\}^{D\times P}$ is computed by maximizing the correlation between the aligned dimensions of $X$ and $S$:
\begin{displaymath}
\text{QVEC} = \max_{\sum_{j}a_{ij}\leq 1}\sum_{i=1}^{D}\sum_{j=1}^{P}r(x_i, s_j)\times a_{ij}
\end{displaymath}
where $x\in \mathbb{R}^{D\times 1}$ denotes a distributional word vector and $s\in  \mathbb{R}^{P\times 1}$ denotes a linguistic word vector. $a_{ij}=1$ iff $x_i$ is aligned to $s_j$, otherwise $a_{ij}=0$. $r(x_i, s_j)$ is the Pearson's correlation between $x_i$ and $s_j$. QVEC-CCA~\cite{ammar2016massively} is extended from QVEC by using CCA to measure the correlation between the distributional matrix and the linguistic vector matrix, instead of cumulative dimension-wise correlation.

Using QVEC and QVEC-CCA, we evaluate the quality of multilingual embeddings for both monolingual (English) and multilingual (English, Danish, Italian) settings. 

As shown in Table~\ref{3lang}, our approaches outperform previous approaches in all cases. Specifically, by augmenting word representation with neighboring words in the common semantic space as in Eq.~\eqref{eq:context_proj}, the performance for monolingual QVEC and QVEC-CCA tasks is consistently improved.  In addition, by aligning character-level compositional representations and linguistic property based clusters in the shared semantic space, the multilingual representation quality is further improved.

\begin{table}[t]
\small
\begin{center}
\begin{tabular}{c| c c c c c}
% \hline
      & Amh & Tig  & Uig & Tur & Eng \\ 
      \toprule
Train &  1,506  & 1,585  &  1,711   &  3,404   &   14,029  \\ %\hline
Dev & 167  &   176   & 190  &  378   &  3,250  \\  %\hline
Test &  711   &  440   &  476   &  1,604   &  3,453  \\ 
% \hline
\end{tabular}
\end{center}
\vspace{-3mm}
\caption{Data statistics (\# of Sentences) for name tagging}
\label{nametaggingData}
\vspace{-3mm}
\end{table}

\subsection{Impact of Bilingual Dictionary Size}

In order to show the impact of the size of bilingual lexicons, we use three languages as a case study, and gradually reduce the size of the lexicons for each pair of languages from 40,000 to 10,000 and to 2,000. Both MultiCluster and MultiSkip by default take advantage of identical strings from any pair of languages when they learn the multilingual embeddings. For fair comparison, we thus use MultiCCA as a baseline. Table~\ref{3lang2000} shows the results. We observe that both MultiCCA and CorrNet approaches are sensitive to the size of the bilingual lexicons. Our approach on the other hand could maintain high performance, even when the bilingual lexicons were reduced to 2,000.

\subsection{Low-Resource Name Tagging}

%\heng{justify why you choose name tagging as an extrinsic task: because it cares about linguistic property}

%\heng{bold the numbers you want to highlight}

%\heng{in result summary highlight: Without using any training data for the target low-resource language, our framework achieves up to 34.1\% name tagging F-score by transferring knowledge and resources from other languages. 
%}

%\heng{need to add a data stats,  the number of documents, the number of names, and script description for each language, from lorelei program, etc}

%\heng{add qualitative analysis as you did for chechen name tagging}

%\heng{do two experiments: 1. Amharic and Tigrinya: related language, the same script, even though they are low-resource languages, we can see gains; 2. Uighur, Turkish and English: different script, and high-resource, medium-resource, low-resource; if you have time add the experiments on Russian-Chechen}

%\heng{if you want to show monolingual embedding results still look good then add a separate table in analysis}

We evaluate the quality of multilingual embeddings on a downstream task by using the embeddings as input features. Here, we use low-resource language name tagging as a target task. We experiment with two sets of languages. The first set Amh+Tig consists of Amharic and Tigrinya. Both languages share the same Ge'ez script and descend from the proto-Semitic language family. The other set Eng+Uig+Tur consists of one high-resource language (English), one medium-resource language (Turkish) and one low-resource language (Uighur). It also consists of two distinct language scripts: English and Turkish use Latin script while Uighur uses Arabic script. 

%\heng{somewhere you should mention this data table, and put down ldc catalogs, don't mention lorelei program for blind review}

\begin{table}[t]
\small
\begin{center}
\begin{tabular}{l||c|c|cc}
% \hline
& & \multicolumn{3}{c}{Multilingual} \\ 
% \cmidrule{3-5}
&  Mono-  &  & \multicolumn{2}{c}{CorrNet}  \\ 
&  lingual  & MultiCCA & W  & W+N+Ch+L \\ 
\toprule%\hline
Amh & 52.0 & 50.6  &  52.4 & \textbf{55.8}  \\
Tig & 78.2 & \textbf{78.4}  &  77.9 & 77.6  \\
Uig & \textbf{70.0} & 63.6  & 66.8  & 66.0  \\
Tur & 73.9 & 65.3  & 72.4  & \textbf{75.6}  \\
% \hline
\end{tabular}
\end{center}
\caption{Name tagging result (F-score, \%) using monolingual embedding and multilingual embeddings.}
\label{monoNametagging}
\end{table}

% \begin{table}[!htp]
% \footnotesize
% \begin{center}
% \begin{tabular}{p{0.8cm}<{\centering}|p{0.9cm}<{\centering}|p{0.9cm}<{\centering}|p{0.9cm}<{\centering}|p{0.9cm}<{\centering}|p{0.9cm}<{\centering}}
% \hline
%       & Amh & Tig  & Uig & Tur & Eng \\ \hline
% Train &  3,012  & 1,585  &  1,711   &  3,404   &   14,029  \\ %\hline
% Dev & 334  &   176   & 190  &  378   &  3,250  \\  %\hline
% Test &  1,422   &  440   &  476   &  1,604   &  3,453  \\ \hline
% \end{tabular}
% \end{center}
% \caption{Number of Sentences in Name Tagging Datasets}
% \label{nametaggingData}
% \end{table}

We use LSTM-CRF architecture~\cite{huang2015bidirectional,lample2016neural,ma2016end} for name tagging. Table~\ref{nametaggingData} shows the statistics of training, development, and test sets for each language released by Linguistic Data Consortium (LDC).\footnote{The annotations are from: Amh (LDC2016E87), Tig (LDC2017E27), Uig (LDC2016E70), Tur (LDC2014E115), Eng~\cite{tjong2003introduction} } For each language pair in each language set, we combine the bilingual aligned words extracted from Wiktionary and extracted from monolingual dictionaries based on identical strings.\footnote{We extracted 23,781 pairs of words for Amh and Tig, 16,868 pairs for Eng and Tur, 3,353 pairs for Eng and Uig, and 3,563 pairs for Tur and Uig.} We evaluate the quality from several aspects:%We project each set of languages into a common semantic space and evaluate the quality of our multilingual embeddings from several aspects:

\begin{table}[t]
\footnotesize
\begin{center}
\begin{tabular}{lc||c|cc}
%\hline
 &  &   &  \multicolumn{2}{c}{CorrNet} \\ 
Train & Test & MultiCCA  &  W  & W+N+Ch+L \\ 
\toprule%\hline
Amh & Tig    &  15.5  &  28.3  &  \textbf{31.7}  \\ %\hline
Tig & Amh    &  11.1  &  12.8  &  \textbf{23.3}  \\ %\hline
\midrule
Eng & Uig &  8.4  &  \textbf{16.9}  &  15.4  \\ %\hline
Tur & Uig &  1.1  &  18.1  &  \textbf{25.6}  \\ %\hline
Eng+Tur & Uig & 8.0 &  20.3  &  \textbf{20.6}  \\ %\hline
\midrule
Eng & Tur  &  20.6 &  \textbf{21.4}  &  17.3  \\ %\hline
Uig & Tur  &  10.4 &  10.1  &  \textbf{17.7}  \\ %\hline
Eng+Uig & Tur &  18.5 &  21.1  &  \textbf{29.4}  \\ 
% \hline
\end{tabular}
\end{center}
\caption{Name tagging performance (F-score, \%) when the tagger was trained on a source language and tested on a target language. CorrNet W+N+Ch+L is the proposed approach with all the cluster types.}
\label{nametagging1}
\vspace{-3mm}
\end{table}

\paragraph{Monolingual embedding quality evaluation} 

Table~\ref{monoNametagging} shows the name tagging performance for each language using the original monolingual embeddings and multilingual embeddings. 
%We can see that, for 
For both Amharic and Turkish, the multilingual embeddings learned from our approach significantly improve over the monolingual embeddings, compared to MultiCCA. In the case of Uighur, all the multilingual embeddings fail to outperform the original monolingual embeddings. We conjecture that this is due to the use of Arabic script in Uighur, which differs from Turkish and English.

\paragraph{Cross-lingual direct transfer}

We further demonstrate the effectiveness of our multilingual embeddings on direct knowledge transfer. In this setting, we train a name tagger on one or two languages using multilingual embeddings and test it on a new language without any annotated data. Table~\ref{nametagging1} shows the performance. %We can see that, for 
For each testing language, our approach achieves better performance than MultiCCA and CorrNet. 
%provides better cross-lingual transfer for name tagging using the multilingual embeddings. 
The closer that the languages are, such as Amharic and Tigrinya, and Turkish and Uighur, the better performance could be achieved, even when they may have distinct language scripts (e.g., Turkish and Uighur). 

We however also notice that a larger extra annotation from another language does not necessarily result in the improvement. For instance, the proposed approach (CorrNet W+N+Ch+L) suffers from English annotated examples when tested on Turkish. This suggests that we need to be careful and aware of linguistic properties among different languages for transfer learning.

% \begin{table}[!htp]
% \footnotesize
% \begin{center}
% \begin{tabular}{p{1.0cm}p{0.7cm}<{\centering}||p{1.2cm}<{\centering}|p{1.2cm}<{\centering}|p{1.5cm}<{\centering}}
% \hline
% Train & Test & MultiCCA  &  CorrNets$^w$  & Our Approach \\ \hline
% Amh & Tig    &  16.1  &  26.8  &  \textbf{32.7}  \\ %\hline
% Tig & Amh    &  10.4  &  13.7  &  \textbf{21.3}  \\ %\hline
% \hline
% Eng & Uig &  8.4  &  \textbf{16.9}  &  15.4  \\ %\hline
% Tur & Uig &  1.1  &  18.1  &  \textbf{25.6}  \\ %\hline
% Eng+Tur & Uig & 8.0 &  20.3  &  \textbf{20.6}  \\ %\hline
% \hline
% Eng & Tur  &  20.6 &  \textbf{21.4}  &  17.3  \\ %\hline
% Uig & Tur  &  10.4 &  10.1  &  \textbf{17.7}  \\ %\hline
% Eng+Uig & Tur &  18.5 &  21.1  &  \textbf{29.4}  \\ \hline
% \end{tabular}
% \end{center}
% \caption{Direct Transfer: Name Tagging Performance (F-score, \%) on Two Sets of Languages: (Amharic, Tigrinya), (English, Turkish, Uyghur)}
% \label{nametagging1}
% \end{table}

\paragraph{Mutual enhancement} 

We finally show the improvement by adding more cross-lingual annotated data and using multilingual embeddings in Table~\ref{nametagging2}. The multilingual embeddings learned by our approach consistently outperforms MultiCCA. More specifically, when there are not enough annotated examples, the performance could be improved by incorporating annotated examples from other languages. This is evident for Amharic, Tigrinya and Uighur. 

% However, for some languages, such as Amharic and Turkish, when their monolingual annotation is sufficient, it may introduce more noise if we directly add annotated data from other languages into training.

\begin{table}[t]
\footnotesize
\begin{center}
\begin{tabular}{lc||c|cc}
%\hline
 &  &   & \multicolumn{2}{c}{CorrNet} \\ 
Train & Test & MultiCCA  &   W  & W+N+Ch+L \\ 
\toprule%\hline
% Amh & Amh      &  50.6 & 52.4   &  \textbf{55.8} \\ %\hline
Tig+Amh & Amh  &  52.9 &  52.1  &  \textbf{56.5}  \\ \midrule
% Tig & Tig      &  78.4 &  \textbf{77.9}  &  77.6 \\ %\hline
Amh+Tig & Tig  &  78.0 &  \textbf{78.1}  &  \textbf{78.7} \\ %\hline
\midrule %\hline
% Uig & Uig      &  63.6  &  \textbf{66.8}  &  66.0 \\ %\hline
Eng+Uig & Uig  &  67.9  &  67.8  &  \textbf{68.3} \\ %\hline
Tur+Uig & Uig  &  67.7  &  67.5  &  \textbf{68.8} \\ %\hline
Eng+Tur+Uig & Uig &  \textbf{68.7}  &  67.4  &  65.9\\ %\hline
\midrule
% Tur & Tur      &  65.3 &  72.4  &  \textbf{75.6} \\ %\hline
Uig-Tur & Tur  &  65.9 &  69.2  &  \textbf{72.8} \\ %\hline
Eng-Tur & Tur  &  66.9 &  70.4  &  \textbf{73.4} \\ %\hline
Eng+Uig+Tur & Tur  &  67.5  &  68.5  & \textbf{72.9} \\ 
% \hline
\end{tabular}
\end{center}
\caption{Name tagging performance (F-score, \%) when the training set for the tagger was enhanced by annotated examples in other languages. CorrNet W+N+Ch+L is the proposed approach with all the cluster types.}
\label{nametagging2}
\vspace{-3mm}
\end{table}

\section{Conclusions and Future Work}
\label{sec:conclusion}
%\heng{let me know when you finish this section}

%an ambitious idea for a unified
% semantic representation that spans thousands of languages. 

% At its core, MCSS is a semantic
% interlingua for representing like concepts/word senses across languages (including English), and is generated using program provided parallel data and harvested monolingual and comparable corpora/resources (e.g., WikiData, Wiktionary). Together, MT and MCSS serve as a powerful crosslanguage semantic bridge. previous work used english as bridge
%Extrinsic evaluations show our framework is very effective at cross-lingual transfer, especially for low-resource language processing tasks. 

%Build a common semantic space across thousands of languages for resource sharing and richer semantic continuous representation for words, concepts

We construct a common semantic space for multiple languages based on a cluster-consistent correlational neural network. It combines word-level alignment and multi-level cluster alignment, including neighbor based clusters, character-level compositional word representations, and linguistic property based clusters induced from the readily available language-universal linguistic knowledge bases. By introducing cluster consistency into multilingual embedding learning, our approach achieved significantly higher performance than state-of-the-art multilingual embedding learning methods through both intrinsic and extrinsic evaluations.  
%consistently higher correlation on QVEC tasks than state-of-the-art multilingual embedding learning methods. Our approach also achieved up to 24.5\% absolute F-score gain over the state of the art on low-resource language name tagging. 
In the future, we will further extend our approach to multi-lingual multi-media common semantic space construction.

%\clearpage
%\input{notes}

% \section*{Acknowledgments}

% The acknowledgments should go immediately before the references.  Do not number the acknowledgments section ({\em i.e.}, use \verb|\section*| instead of \verb|\section|). Do not include this section when submitting your paper for review.

%\bibliography{acl2018}
\bibliography{acl2018}
\bibliographystyle{acl_natbib}

% \appendix

% \section{Supplemental Material}
% \label{sec:supplemental}

\end{document}